\pdfoutput=1

\documentclass[11pt]{article}

\usepackage{EACL2023}

\usepackage{times}
\usepackage{latexsym}
\usepackage{graphicx}
\usepackage{microtype}
\usepackage{multirow}
\usepackage{inconsolata}
\usepackage{pgfplots}
\pgfplotsset{compat=1.17}
\usepackage{enumitem}
\usepackage[T1]{fontenc}

\usepackage[utf8]{inputenc}

\usepackage{microtype}

\usepackage{inconsolata}

\newcommand\blfootnote[1]{%
  \begingroup
  \renewcommand\thefootnote{}\footnote{#1}%
  \addtocounter{footnote}{-1}%
  \endgroup
}

\title{MasonPerplexity at ClimateActivism 2024: Integrating Advanced Ensemble Techniques and Data Augmentation for Climate Activism Stance and Hate Event Identification}

 \author{Al Nahian Bin Emran\textsuperscript{*}, Amrita Ganguly\textsuperscript{*}, Sadiya Sayara Chowdhury Puspo \\ {\bf Dhiman Goswami, Md Nishat Raihan} \\ George Mason University, USA \\
 \texttt{\{abinemra, agangul\}@gmu.edu}}

\begin{document}
\maketitle
\begin{abstract}
The task of identifying public opinions on social media, particularly regarding climate activism and the detection of hate events, has emerged as a critical area of research in our rapidly changing world. With a growing number of people voicing either to support or oppose to climate-related issues - understanding these diverse viewpoints has become increasingly vital. Our team, \textit{MasonPerplexity}, participates in a significant research initiative focused on this subject. We extensively test various models and methods, discovering that our most effective results are achieved through ensemble modeling, enhanced by data augmentation techniques like back-translation. In the specific components of this research task, our team achieved notable positions, ranking $5^{th}$, $1^{st}$, and $6^{th}$ in the respective sub-tasks, thereby illustrating the effectiveness of our approach in this important field of study. \blfootnote{\bf * denotes Equal Contribution}
\end{abstract}

\section{Introduction}

In the ever-evolving landscape of climate change activism, encouraging meaningful conversations and comprehending how things change throughout events depends critically on the ability to recognize hate speech and the understanding of attitude during these events. This paper presents our effort in the Shared Task on Hate Speech and Stance Detection during Climate Activism \cite{thapa2024stance}, where our goal is to develop effective models for hate speech detection, target identification, and stance detection.

This task consists of three subtasks that work together to support an integrated approach to event identification. The goal of the first subtask is to identify whether the given text contains hate speech or not. The second subtask focuses on identifying if people, groups, or communities are targets of hate speech. Lastly, Stance Detection provides insight into the dynamics of climate activism protests by assessing the support, opposition, or neutrality indicated within texts.

Our paper serves as a comprehensive system description, outlining the approaches and models used to address these subtasks within the framework of activist events related to climate change. We present our ensemble method for identifying hate speech, which combines robust models like XLM-roBERTa-large \cite{xlmr}, BERTweet-large \cite{berttweetner1}, and fBERT \cite{fbert}. Notably, for Target Detection, the best-performing model is BERTweet-large \cite{ushio-camacho-collados-2021-ner} while BERTweet-base \cite{vinaibertweet} excels in Stance Detection.

We also discuss our fine-tuning strategies and dataset augmentation techniques, demonstrating our commitment to refining model performance. Our approach's effectiveness is demonstrated by our remarkable F1 scores of 0.8885, 0.7858, and 0.7373. Furthermore, our team named \textit{MasonPerplexity} has secured $5^{th}$, $1^{st}$, and $6^{th}$ ranks in the respective subtasks, underscoring the competence of our models in comparison to peers.

Through this paper, we aim to contribute to the advancement of hate speech and stance detection in the context of climate activism, fostering a safer and more informed space for dialogue and understanding during crucial events. We employ ensemble methods to better classify the texts - our approach increases the accuracy metrics for the first sub-task where we encounter a comparatively larger amount of data. We also use data augmentation methods, which further improve our results.

\section{Related Works}
The paper \cite{hsdnlp-parihar-2021} explores the rising concern of hate speech on the internet and its potential impact. It emphasizes machine learning and deep learning models in automatically identifying hate speech. In \cite{DBLP:journals/corr/abs-1712-06427}, English tweets are subjected to supervised classification using n-gram features and a linear SVM classifier. Even at 78\% accuracy, it is still difficult to discern offensive language from hate speech. By combining recurrent neural networks and user features, \cite{Pitsilis2018EffectiveHD} outperforms current systems and achieves a remarkable F-score of 0.9320 on a Twitter dataset. Additionally, Warner and Hirschberg \cite{inproceedings} define hate speech and despite limitations in capturing larger language patterns, SVM classification can detect anti-Semitic speech with 94\% accuracy.

The literature on target detection in hate speech unveils valuable insights through various studies in the field. \cite{lemmens-etal-2021-improving2} focuses on Dutch Facebook comments, exploring hateful metaphors to enhance hate speech type and target detection. The study incorporates manual metaphor annotations as features for SVM and BERT models, observing improvements in F1 scores. Conversely, \cite{zampieri-etal-2019-predicting2} proposes a hierarchical annotation scheme for offensive language in English tweets, creating the OLID dataset. The study employs SVM, CNN, and BiLSTM models, achieving notable results and providing a valuable resource for offensive language research.

Stance detection, a crucial aspect of NLP, involves determining a person's position towards a concept. \cite{10.1145/3404835.3462815} outlines the significance and challenges in this domain, emphasizing its relation to sentiment analysis, emotion detection, and other tasks. It highlights the evolution facilitated by shared tasks, varied approaches, including traditional SVMs and newer LSTM models, and the necessity of annotated datasets. Additionally, \cite{upadhyaya-etal-2023-toxicity} introduces a multitasking approach, enhancing performance on multiple datasets, and showcasing the potential of incorporating auxiliary tasks. Furthermore, \cite{DBLP:journals/corr/abs-1803-08910} contributes a valuable stance-annotated Turkish Twitter dataset, showcasing the diversity of research efforts in stance detection.

\section{Datasets}
From the tables for Subtask \ref{tab:ST1}, Subtask \ref{tab:ST2}, and Subtask \ref{tab:ST3}, it is evident that the dataset \cite{shiwakoti2024analyzing} is imbalanced across different labels.

\subsection{Hate Speech Detection}
In subtask A, the distribution between NON-HATE and HATE is heavily skewed towards NON-HATE, with approximately 87.66\% in the training set, 87.83\% in the evaluation set, and 87.96\% in the test set. This indicates a significant class imbalance, which may pose challenges for model training and evaluation.

\subsection{Target Detection}
In subtask B, there is an imbalance among the labels INDIVIDUAL, ORGANIZATION, and COMMUNITY. The majority of instances belong to the individual category, with around 80.54\% in the training set, 80.00\% in the evaluation set, and 80.67\% in the test set.

\subsection{Stance Detection}
Subtask C exhibits an imbalance, between the SUPPORT, OPPOSE, and NEUTRAL labels. SUPPORT dominates the dataset, comprising 59.42\% in the training set, 57.46\% in the evaluation set, and 58.96\% in the test set, where OPPOSE has respective percentages 9.61\%, 9.80\%, and 9.03\%.

In summary, the dataset for all three subtasks is not well-balanced, and addressing this imbalance may be crucial for developing models that generalize well across different classes.

\begin{table}[h]
\centering
\begin{tabular}{lccc}
\hline
\multicolumn{4}{c}{\textbf{Subtask A}} \\
\hline
\textbf{Label} & \textbf{Train} & \textbf{Eval} & \textbf{Test } \\
\hline
\textsc{non-hate} & 87.66 & 87.83 & 87.96 \\
\textsc{hate} & 12.34 & 12.17 & 12.04 \\
\hline
\end{tabular}
\caption{label wise data percentage of subtask A}
\label{tab:ST1}
\end{table}

\begin{table}[h]
\centering
\begin{tabular}{lccc}
\hline
\multicolumn{4}{c}{\textbf{Subtask B}} \\
\hline
\textbf{Label} & \textbf{Train} & \textbf{Eval} & \textbf{Test} \\
\hline
\textsc{individual} & 80.54 & 80.00 & 80.67 \\
\textsc{organization} & 15.02 & 15.33 & 15.33 \\
\textsc{community} & 4.44 & 4.67 & 4.00 \\
\hline
\end{tabular}
\caption{label wise data percentage of subtask B}
\label{tab:ST2}
\end{table}

\begin{table}[h]
\centering
\begin{tabular}{lccc}
\hline
\multicolumn{4}{c}{\textbf{Subtask C}} \\
\hline
\textbf{Label} & \textbf{Train} & \textbf{Eval} & \textbf{Test} \\
\hline
\textsc{support} & 59.42 & 57.46 & 58.96 \\
\textsc{oppose} & 9.61 & 9.80 & 9.03 \\
\textsc{neutral} & 30.97 & 32.74 & 32.01 \\
\hline
\end{tabular}
\caption{label wise data percentage of subtask C}
\label{tab:ST3}
\end{table}

\section{Experiments}
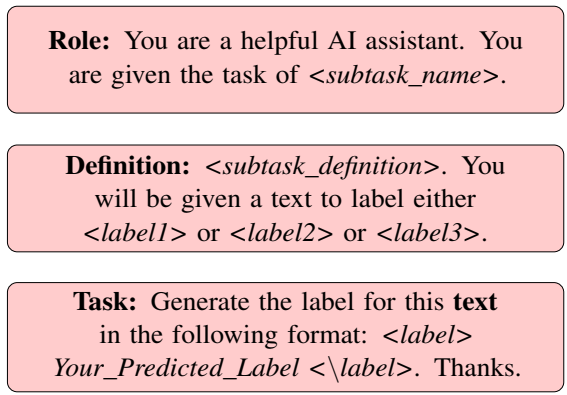
\begin{figure}[h]
\centering
\scalebox{.92}{
\begin{tikzpicture}[node distance=1cm]
    \tikzstyle{block} = [rectangle, draw, fill=red!20, text width=\linewidth, text centered, rounded corners, minimum height=4em]
    \tikzstyle{operation} = [text centered, minimum height=1em]
    \node [block] (rect1) {\textbf{Role:}{ You are a helpful AI assistant. You are given the task of \textit{<subtask\_name>}. }};
    \node [operation, below of=rect1] (plus1) {};
    \node [block, below of=plus1] (rect2) {\textbf{Definition:}{ \textit{<subtask\_definition>}. You will be given a text to label either \textit{<label1>} or \textit{<label2>} or \textit{<label3>}. }};
    \node [operation, below of=rect2] (plus2) {};
    \node [block, below of=plus2] (rect3) {\textbf{Task:}{ Generate the label for this \textbf{text} in the following format: \textit{<label> Your\_Predicted\_Label <$\backslash$label>}. Thanks.}};
\end{tikzpicture}
}
\caption{Sample GPT-3.5 prompt.}
\label{fig:prompt1}
\end{figure}

In subtask A, we initially employ GPT3.5 \cite{openai2023gpt35turbo} zero shot and few shot prompting with Test F1 score 0.66 and 0.73. The prompt provided to GPT3.5 is available in Figure \ref{fig:prompt1}.

Then we use  BERTweet large \cite{berttweetner1} \cite{berttweetner2}, XLM-R \cite{xlmr}, HATE-BERT \cite{caselli-etal-2021-hatebert} and fBERT \cite{fbert}. Following this, we adopt a weighted ensemble approach (Ensemble 1) for the best three models (BERTweet, XLM-R, fBERT). Similarly, we perform another weighted ensemble approach (Ensemble 2) with the same models only replacing BERTweet with HATE-BERT, as these two models show the same F1 score on test data with the same setting. However, the former ensemble strategy yields the highest F1 score for this task. 

To address class imbalance in subtask A, we implement back translation by converting the training data of those specific labels that have a smaller ratio with respect to the whole training set through various languages, including Xosha to Twi to English, Lao to Pashto to Yoruba to English, Yoruba to Somali to Kinyarwanda to English, and Zulu to Oromo to Shona to Tsonga to English. This approach significantly contributed to improving the overall F1 score of Ensemble 1 from 0.85 to 0.88 and Ensemble 2 from 0.86 to 0.89.

We follow the approach of back translation of \cite{raihan-etal-2023-nlpbdpatriots}. For this, we select languages that demonstrate limited or no cultural overlap with the original language featured in the dataset. Xosha, Twi, Lao, Pashto, Yoruba, Somali, Kinyarwanda, Zulu, Oromo, Shona, and Tsonga are languages that are very diverse culturally and geographically. This diversity underscores the significance of considering a wide range of cultural and geographical influences when working with these languages.
By intentionally selecting these languages without cultural overlap, we introduce a purposeful aspect of diversity, mitigating potential biases, and enhancing the dataset with a broader spectrum of linguistic expressions. Moreover, the Ensemble method with majority voting is also proven helpful in this type of case where a single model may not label the data correctly due to class imbalance \cite{goswami-etal-2023-nlpbdpatriots}. For instance, when two out of three models predict a sentence as a hate event, the sentence is subsequently labeled as a hate event through the application of majority voting.

In subtask B, we utilize BERTweet-large \cite{berttweetner1}, BERT base \cite{bertbaseuncase}, and XLM-R \cite{xlmr}. Additionally, like subtask 1, we implement back translation using the same language sequences mentioned earlier to address class imbalance. Notably, BERTweet large \cite{berttweetner1} demonstrates the highest F1- score among these models. We also use GPT3.5 zero shot and few shot prompting with 0.63 and 0.64 test F1 scores.

BERTweet-large \cite{berttweetner1}, BERT base \cite{bertbaseuncase}, and BERTweet base \cite{vinaibertweet} models are applied in subtask C for stance detection. Among these models, the BERTweet base achieves the highest F1 score. F1 score for GPT3.5 zero shots and few shot prompting are 0.63 and 0.67.

Hyperparameters of all the models used excluding GPT3.5 in the experiments are available in Figure \ref{tab:training_config}.

\begin{table}[ht]
\centering
\begin{tabular}{lc}
\hline
\textbf{Parameter} & \textbf{Value} \\
\hline
Learning Rate & \(1e-5\) \\
Train Batch Size & 8 \\
Test Batch Size & 8 \\
Epochs & 5 \\
Dropout & 0.2\\
\hline
\end{tabular}
\caption{Training Configuration Parameters}
\label{tab:training_config}
\end{table}

\section{Results}
The results in Tables \ref{tab:R1}, \ref{tab:R2}, and \ref{tab:R3} provide a comprehensive evaluation of various NLP models across the three subtasks of the shared task. 

In subtask A, our ensemble approach (Ensemble 2 with HATE-BERT, XLM-R and fBERT models) secures the fifth rank. For subtask B, BERTweet large secures the top position (Rank 1), while in subtask C, we achieve the sixth rank utilizing BERT-Base.

\begin{table}[h]

\centering
\resizebox{\linewidth}{!}{
\begin{tabular}{lcc}
\hline
\textbf{Model} & \textbf{Eval F1} & \textbf{Test F1} \\
\hline
\textsc{GPT3.5-(Zero Shot)} & -- & 0.66\\
\textsc{GPT3.5-(Few Shot)} & -- & 0.73\\
\hline
\textsc{hate-BERT} & 0.88 & 0.83\\
\textsc{BERTweet-large} & 0.89 & 0.84\\
\textsc{XLM-R} & 0.89 & 0.85 \\
\textsc{F-BERT} & 0.90 & 0.85\\
\hline
\textsc{*Ensemble 1} & 0.90 & 0.85 \\
\hline
\textsc{**Ensemble 2} & 0.91 & 0.86\\
\hline
\textsc{hate-BERT (AUG.)} & 0.91 & 0.87\\
\textsc{BERTweet-large (AUG.)} & 0.92 & 0.87\\
\textsc{XLM-R (AUG.)} & 0.91 & 0.88 \\
\textsc{F-BERT (AUG.)} & 0.93 & 0.88\\
\hline
\textsc{*Ensemble 1 (AUG.)} & 0.93 & 0.88 \\
\hline
\textsc{**Ensemble 2 (AUG.)} & 0.94 & \textbf{0.89} \\
\hline
\end{tabular}
}
\caption{Results of subtask A (before and after data augmentation). *Ensemble 1 (BERTweet-large, XLM-R, fBERT), **Ensemble 2 (HATE-BERT, XLM-R, fBERT)}
\label{tab:R1}
\end{table}

\begin{table}[h]

\centering
\begin{tabular}{lcc}
\hline
\textbf{Model} & \textbf{Eval F1} & \textbf{Test F1} \\
\hline
\textsc{GPT3.5-(Zero Shot)} & -- & 0.63\\
\textsc{GPT3.5-(Few Shot)} & -- & 0.64\\
\hline
\textsc{XLM-R} & 0.75 & 0.60\\
\textsc{BERT-base} & 0.86 & 0.69 \\
\textsc{BERTweet-large} & 0.97 & \textbf{0.79}\\

\hline
\end{tabular}
\caption{Results of subtask B.}
\label{tab:R2}
\end{table}

\begin{table}[h]

\centering
\begin{tabular}{lcc}
\hline
\textbf{Model} & \textbf{Eval F1} & \textbf{TEST F1} \\
\hline
\textsc{GPT3.5-(Zero Shot)} & -- & 0.63\\
\textsc{GPT3.5-(Few Shot)} & -- & 0.67\\
\hline
\textsc{BERT-base} & 0.71 & 0.69 \\
\textsc{BERTweet-large} & 0.71 & 0.70\\
\textsc{BERTweet-base} & 0.80 & \textbf{0.74} \\

\hline
\end{tabular}
\caption{Results of subtask C.}
\label{tab:R3}
\end{table}

\section{Error Analysis}
Upon evaluating our models' performance across the three subtasks, we identify several key sources of errors that contributed to limiting our scores.

In subtask A on hate speech detection, our ensemble model struggles with longer text segments that express hate in subtle or nuanced ways. The models are not always able to pick up on the underlying mocking or criticism woven into complex rhetorical devices. Additionally, sarcasm and irony continue to pose challenges, as models interpret literally what is meant to convey the opposite meaning.

For subtask B on target identification, errors frequently occur in distinguishing between organizations and communities as categories. Our models have difficulty consistently applying the definitions and criteria that delineate these two groups as targets of hate speech. There are also inconsistencies in labeling individual people who are associated with or represent a broader community.

Regarding subtask C on stance detection, our models struggle to some extent with longer text segments, having more trouble identifying stances from among nuanced discussions. Shorter, more direct statements of opposition or support were simpler for the models to categorize accurately.

\begin{figure} [!h]
  \centering
  \includegraphics[width=\linewidth]{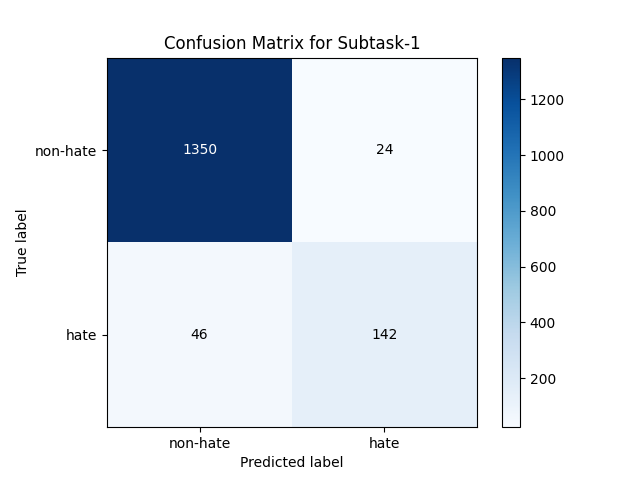}
  \caption{Confusion Matrix for Hate Speech Detection}
  \label{fig:st1}
\end{figure}

\begin{figure} [!h]
  \centering
  \includegraphics[width=\linewidth]{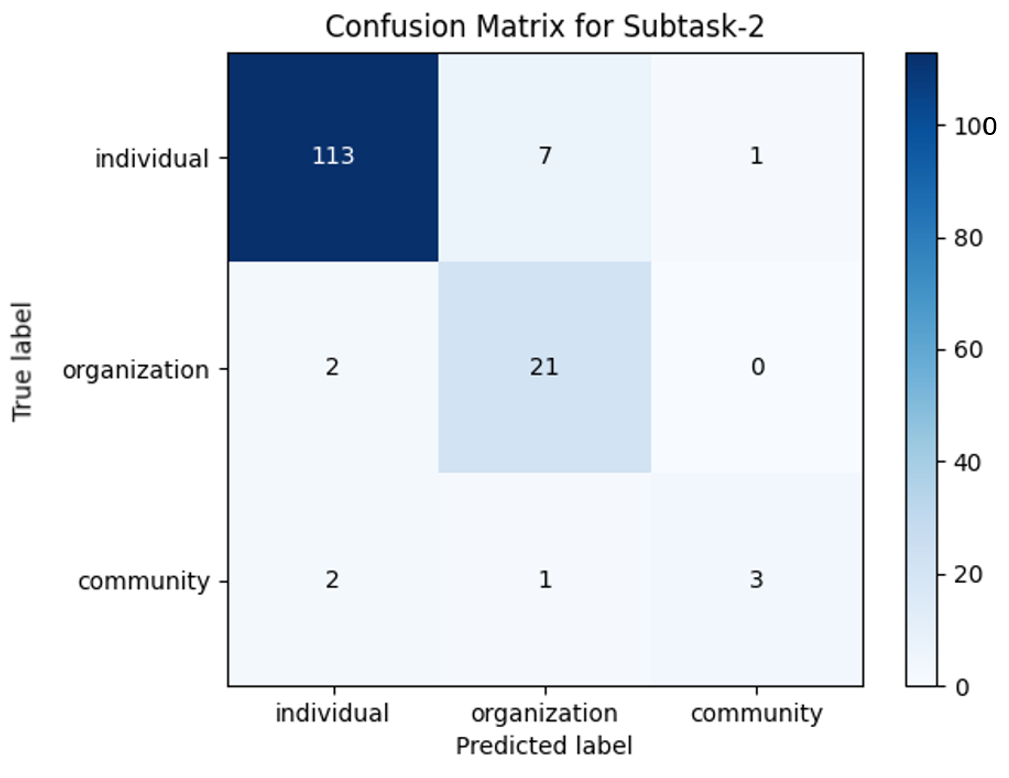}
  \caption{Confusion Matrix for Target Detection}
  \label{fig:st2}
\end{figure}

\begin{figure} [!h]
  \centering
  \includegraphics[width=\linewidth]{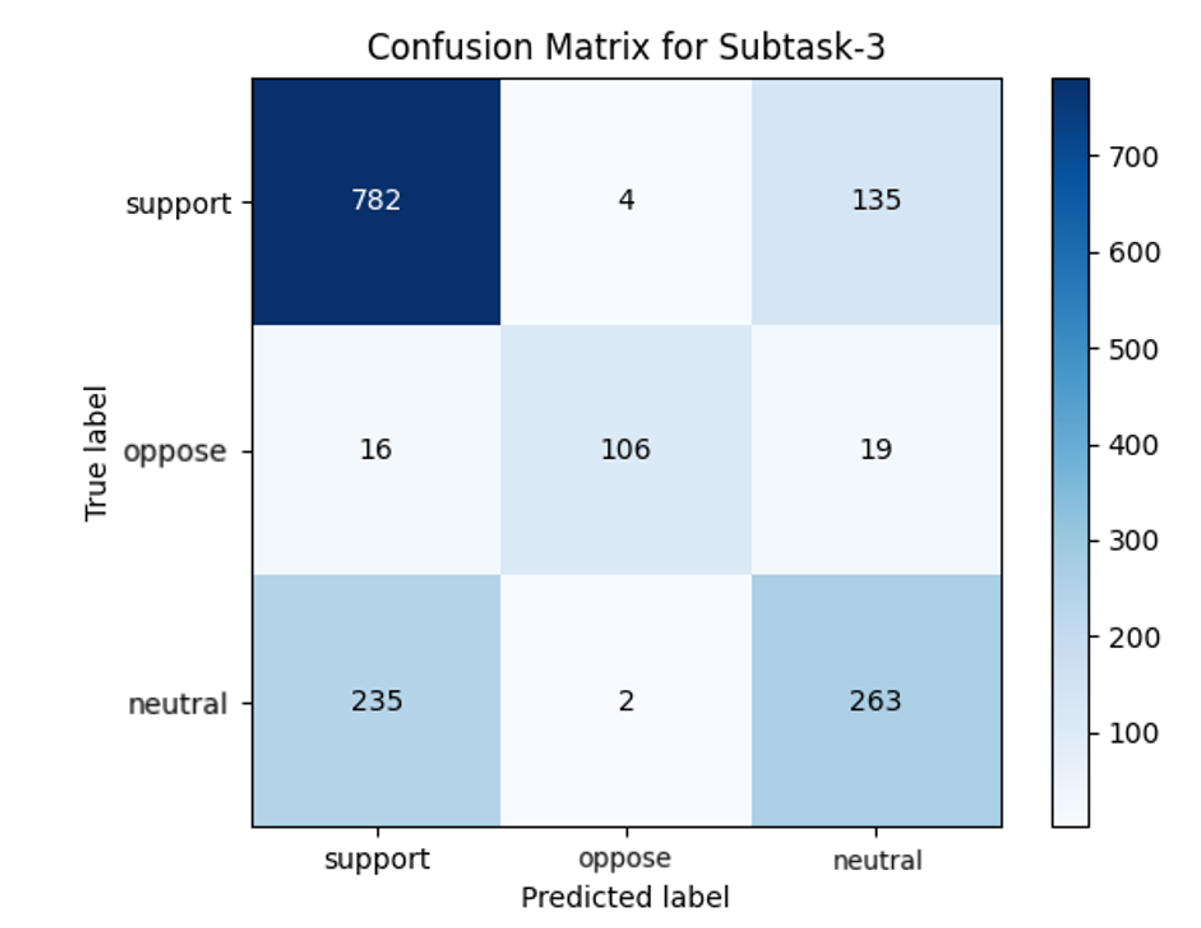}
  \caption{Confusion Matrix for Stance Detection}
  \label{fig:st3}
\end{figure}

To visualize label-wise models' performance we can see the Figures \ref{fig:st1}, \ref{fig:st2}, and \ref{fig:st3} of confusion matrices for all the subtasks. 

\section{Conclusion}
In conclusion, the \textit{MasonPerplexity} team has made significant strides in the domain of detecting climate activism stances and hate events on social media. Through a comprehensive evaluation of various models, our research underscores the efficacy of ensemble modeling coupled with data augmentation techniques like back-translation. Our achievements in the shared task, marked by rankings of $5^{th}$, $1^{st}$, and $6^{th}$ in the respective subtasks, reflect the potential of our methodologies in addressing the complexities of sentiment analysis in the context of climate activism. 

There are several avenues for future research. Firstly, addressing the challenge of label imbalance in our dataset could enhance the accuracy and reliability of our models. Exploring advanced techniques in data sampling or synthetic data generation may provide viable solutions. Secondly, the refinement of label quality through more rigorous annotation processes or leveraging semi-supervised learning techniques could further improve model performance. Finally, the integration of Large Language Model (LLM) fine-tuning presents a promising direction. Fine-tuning pre-trained models specifically for the nuances of climate activism discourse and hate speech detection could yield more nuanced and contextually aware results. Additionally, expanding our research to include multilingual datasets would enhance the applicability and relevance of our findings in a global context, fostering a more comprehensive understanding of public sentiment on climate issues worldwide.

\section*{Limitations}
This study encounters some limitations that affect its outcomes. The first issue is with the balance of labels in our dataset. We have more examples of some types of data than others, a problem known as label imbalance. This imbalance can lead our model to be better at recognizing the more common types and not as good with the rare ones, creating a bias in our results. The second limitation is the quality of the labels themselves. In our dataset, some labels are incorrect or not consistent with each other. This poor quality can confuse the model, making it harder for it to learn correctly and possibly leading to inaccurate results. Lastly, we did not fine-tune Large Language Models (LLMs) for our specific task. Fine-tuning is a process where a pre-trained model, like an LLM, is further trained on a specific type of data. Not doing this fine-tuning means we may not be taking full advantage of the LLM's capabilities, which can improve our model's understanding of complex patterns in climate activism and hate event data. However, due to a lack of computing resources, we are not fine-tuning.

\section*{Acknowledgment}

We would like to thank the shared task organizers for providing us with the dataset used in our study. Moreover, we also want to express our gratitude to Dr. Marcos Zampieri\footnote{\url{https://www.gmu.edu/profiles/mzampier}} for his effective guidelines throughout the span of the competition.

\section*{Ethics Statement}
The present study, which centers on the identification of Climate Activism Stance and Hate Event, rigorously adheres to the \href{https://www.aclweb.org/portal/content/acl-code-ethics}{ACL Ethics Policy} and seeks to make a valuable contribution to the realm of online safety. The dataset is supplied to us by the organizers and has undergone anonymization to secure the privacy of the users. The technology in question possesses the potential to serve as a beneficial instrument for the moderation of online content, thereby facilitating the creation of safer digital environments. However, it is imperative to exercise caution and implement stringent regulations to prevent its potential misuse for purposes such as monitoring or censorship.

\bibliography{anthology,custom}
\bibliographystyle{acl_natbib}

\end{document}